\documentclass[conference]{IEEEtran}

\usepackage{url}
\usepackage{booktabs}
\usepackage{colortbl}
\usepackage{hyperref}
\usepackage{arydshln}
\usepackage{amsmath}
\usepackage[nameinlink,capitalise]{cleveref}
\usepackage{multirow}
\definecolor{newcolor}{rgb}{.8,.349,.1}
\usepackage{graphicx}
\usepackage{authblk}
\usepackage{amsmath,amssymb,amsfonts}
\usepackage{algorithmic}
\usepackage{rotating}
\usepackage{multirow}
\usepackage{makecell}
\usepackage{hhline}
\usepackage{graphicx}
\usepackage{textcomp}
\usepackage{color}
\usepackage{booktabs}
\usepackage{hyperref}
\usepackage{cleveref}
\usepackage{soul}

\begin{document}
%
\title{Self-Path: Self-supervision for Classification of Pathology Images with Limited Annotations}


\author[1,2]{Navid Alemi Koohbanani}
\author[3]{Balagopal Unnikrishnan}
\author[4]{Syed Ali Khurram}
\author[3]{Pavitra Krishnaswamy}
\author[1,2]{, and Nasir Rajpoot}

\affil[1]{Department of Computer Science, University of Warwick, UK}
\affil[2]{Alan Turing Institute, UK}
\affil[3]{Department of Machine Intellection,  Institute for Infocomm Research, Singapore}
\affil[4]{School of Clinical Dentistry, University of Sheffield, UK}

\maketitle


\begin{abstract}
While high-resolution pathology images lend themselves well to `data hungry' deep learning algorithms, obtaining exhaustive annotations on these images is a major challenge. In this paper, we propose a self-supervised CNN approach to leverage unlabeled data for learning generalizable and domain invariant representations in pathology images. The proposed approach, which we term as Self-Path, is a multi-task learning approach where the main task is tissue classification and pretext tasks are a variety of self-supervised tasks with labels inherent to the input data. We introduce novel domain specific self-supervision tasks that leverage contextual, multi-resolution and semantic features in pathology images for semi-supervised learning and domain adaptation. We investigate the effectiveness of Self-Path on 3 different pathology datasets. Our results show that Self-Path with the domain-specific pretext tasks achieves state-of-the-art performance for semi-supervised learning when small amounts of labeled data are available. Further, we show that Self-Path improves domain adaptation for classification of histology image patches when there is no labeled data available for the target domain. This approach can potentially be employed for other applications in computational pathology, where annotation budget is often limited or large amount of unlabeled image data is available.
\end{abstract}


\section{Introduction}
\label{sec:introduction}




The recent surge in the area of computational pathology can be attributed to the increasing ubiquity of digital slide scanners and the consequent rapid rise in the amount of raw pixel data acquired by scanning of histology slides into digital whole-slide images (WSIs). These developments make the area of computational pathology ripe ground for deep neural network (DNN) models. In recent years, there have been notable successes in training DNNs for pathology image analysis and automated diagnosis of disease in the histopathology domain \cite{skrede2020deep}. The performance and generalizability of most DNNs is, however, highly dependent on the availability of large and diverse amounts of annotated data. Although the use of digital slide scanners have made large amounts of raw data available, development of DNN based algorithms remains bottlenecked by the need for extensive annotations on diverse datasets. 

In pathology, annotation burden can pose a large problem -- even more so when compared to natural scene images. WSIs are by nature high resolution images (sometimes with slide dimensions as large as 200,000 $\times$ 150,000 pixels) -- this hinders exhaustive annotations. For even simple use cases like detecting tumor regions or isolated tumor cells in WSIs, pathologists annotating the data need to look at regions of the tissue at multiple levels of magnification. So, even simple labeling of regions of interest can be quite demanding. This issue is compounded by the fact that the whole image can only be annotated part by part owing to its large size. Further, the annotation effort requires expert domain knowledge and significant investment on the part of specialized pathologists. To overcome these challenges, when training DNNs on new pathology image datasets, it would be desirable to pursue one or both of the following strategies: (a) labeling small amounts of the new dataset and making use of the larger pool of the unlabeled data, and/or (b) using existing labeled datasets which closely match the new dataset. 

For strategy (a), \emph{semi-supervised deep learning} approaches that learn with small amounts of labeled data and leverage larger pools of unlabeled data to boost performance can be employed. These approaches have been widely demonstrated in the computer vision community for natural scene images. Particularly popular techniques include Mean Teacher \cite{tarvainen2017mean} and Virtual Adversarial Training (VAT) \cite{miyato2018virtual}. Recently, these approaches have also been applied to the area of computational pathology to address tasks such as clustering \cite{peikari2018cluster}, segmentation \cite{li2018based} and image retrieval \cite{sparks2016out}. However, due to the high dimensionality of the images, the multi-scale nature of the problem, the requirement of contextual information and texture-like nature of sub-patches extracted from slides, the direct translation of popular semi-supervised algorithms into pathology classification tasks is not feasible. 

For strategy (b), \emph{domain adaptation} approaches that transfer knowledge from existing resources for related tasks to the classification task-at-hand can be employed. However, due to variations in tissue, tumor types, and stain appearance during image acquisition, different pathology image datasets appear quite distinct from one another. In addition, for some rare tissue or tumor types, there may be no annotated datasets available for such knowledge transfer. Hence, direct translation of existing domain adaptation algorithms which work for natural vision images may not be possible. Yet, unlabeled data for related tasks are largely available and are less prone to bias \cite{noroozi2016unsupervised}. Hence, when dealing with limited annotations, such unlabeled data can be used to capture the shared knowledge or to learn representations that can improve model performance. 

To address the dual challenges of low annotations and domain adaptation in histopathology, it is possible to use unlabeled data in a self-supervised manner. In this setup, the model is supervised by labels that come inherently from the data itself without any additional manual annotations. These labels can represent distinct morphological, geometrical and contextual content of the images. Models trained on these `free' labels can learn representations that can improve performance for a variety of tasks such as classification, segmentation and detection \cite{jing2020self}. Self-supervision tasks can be used together with the main supervised task in a multi-task setup to improve performance for semi-supervised learning and domain adaptation \cite{zhai2019s4l}. However, self-supervised tasks proposed in the literature so far are mainly based on characteristics of natural scene images, which are very different from histology images. For instance, common self-supervision tasks focus on predicting the degree of rotation, flipping, and/or the relative position of objects. While these are meaningful concepts for natural scene images, they do not carry much relevance for histopathology images. Specifically, while the degree of rotation could help to also learn semantic information present in a natural image, it would not make sense for pathology images because they have no sense of global orientation \cite{graham2020dense}. 


In this paper, we propose the \textbf{Self-Path} framework to leverage self-supervised tasks customized to the requirements of the histopathology domain, and enhance DNN training in scenarios with limited or no annotated data for the task at hand. Our main contributions are summarized as follows:
\begin{itemize}
    \item We introduce a generic and flexible self-supervision based framework, Self-Path, for classification of pathology images in the context of limited or no annotations.
    \item We propose 3 novel pathology domain specific self-supervision tasks aimed at utilizing contextual, multi-resolution and semantic features in histopathology images.
    \item We conduct a detailed investigation on the effect of various self-supervision tasks for semi-supervised learning and domain adaptation for three varied histopathology image classification datasets.
    \item We demonstrate that Self-Path achieves state-of-the art performance for semi-supervised learning in limited annotation regimes where ~1-2$\%$ of the whole dataset is annotated. 
    \item We further show that Self-Path can leverage existing annotated resources on related tasks, enable domain adaptation, and achieve competitive performance when no annotated data is available for the target task.
\end{itemize} 

\subsection{Related Work}

\textbf{Semi-supervised Learning}: Semi-supervised deep learning approaches are widely studied in the computer vision literature \cite{van2020survey}. Popular methods utilize forms of pseudo labelling and consistency regularization, and utilize small amounts of labeled data alongside larger pools of unlabeled data for learning. Pseudo-labeling approaches \cite{lee2013pseudo} use available labels to train a model and impute labels on the unlabeled samples which are in turn used in training. MixMatch extends pseudo-labeilng by adding temperate sharpening along with the mix-up augmentation \cite{berthelot2019mixmatch} . Consistency-based methods regularize the model by ensuring stable outputs for various augmentations of the same sample. These can be done by enforcing consensus between temporal ensembles of network outputs like in Pi-Model \cite{laine2016temporal}, or between perturbed images fed to a network and its EMA averaged counterpart like in Mean Teacher\cite{tarvainen2017mean}. Virtual adversarial training(VAT) \cite{miyato2018virtual} generates the perturbed images in an adversarial fashion to smooth the margin in the direction of maximum vulnerability. These methods ensure generalizability against significant image perturbations, move the margin away from high-density regions, and enable strong performance on benchmark natural scene image tasks with low annotation budgets. 

However, semi-supervised learning has not been sufficiently explored in pathology image analysis. At the time of this writing, only 6 papers investigate semi-supervised learning for the histopathology domain. In \cite{li2018based}, Li et. al proposed an EM-based approach for semi-supervised segmentation of histology images. \cite{peikari2018cluster} proposed a cluster based semi-supervised approach to identify high-density regions in the data space which were then used by supervised SVM in finding the decision boundary. Deep multiple instance learning and contrasting predictive coding were used together in \cite{lu2019semi} to overcome the scarcity of labeled data for breast cancer classification. Jaiswal {\em et al.} \cite{jaiswal2019semi} used pseudolabels for improving the network performance for metastasis detection of breast cancer. Su {\em et al.} \cite{su2019local} employed global and local consistency losses for mean teacher approach for nuclear classification. Shaw et. al \cite{shaw2020teacher} also proposed to use pseudo-labels of unlabeled images for fine-tuning the model iteratively to improve performance for colorectal image classification. Yet, there is scope for improvement to close the gap between fully supervised baselines and semi-supervised methods employing just a few labeled pathology images. 

\textbf{Domain Adaptation}: Domain adaptation methods focus on adapting models trained on a source dataset to perform well on a target dataset. Leading-edge techniques mainly use adversarial training for aligning the feature distributions of different domains. Popular domain-adversarial learning-based methods\cite{tzeng2017adversarial, ganin2016domain} use a domain discriminator to classify the domain of images. These methods play a minimax game where the discriminator is trained to distinguish the features from the source or target sample, while the feature generator is trained to confuse the discriminator. \cite{shen2018wasserstein} employed adversarial learning and minimized Wassertein distance between domains to learn domain-invariant features. Image-translation methods minimize the discrepancy between the two domains at an image-level \cite{bousmalis2017unsupervised}. In pathology, Ren {\em et al.} \cite{ren2018adversarial} employed adversarial training for domain adaptation across acquisition devices (scanners) in a prostate cancer image classification task. \cite{xing2019adversarial} used CycleGAN to translate across domains for a cell/nuclei detection task. \cite{stacke2019closer} introduced a measure for evaluating distance between domains to enhance the ability to identify out-of-distribution samples in a tumor classification task. Yet, most practical domain adaptation techniques require labeling of target domain data, and the applicability of state-of-the-art unsupervised domain adaptation approaches for histopathology is yet to be widely established. 

\textbf{Self-Supervision}: Self-supervision employs pretext tasks (based on annotations that are inherent to the input data) to learn representations that can enhance performance for the downstream task \cite{jing2020self}. Autoencoders \cite{hinton2006reducing} are the simplest self-supervised task, where the goal is to minimize reconstruction error and the proxy labels are the values of image pixels. Other self-supervised tasks in the literature are image generation \cite{jing2020self} , inpainting \cite{pathak2016context}, colorizing grayscale images \cite{larsson2017colorization}, predicting rotation \cite{gidaris2018unsupervised}, solving jigsaw puzzle \cite{noroozi2017representation}, and contrastive predictive coding \cite{henaff2019data}. Although the classical self-supervision approaches requires no additional annotations, it is also possible to leverage small amounts of labeled data within a self-supervision framework. For example, S4L \cite{zhai2019s4l} showed that the pretext task (e.g., rotation, self-supervised exemplar \cite{dosovitskiy2014discriminative}) can benefit from small amount of labeled data alongside larger unlabeled data. As there is no large labeled dataset akin to ImageNet for pretraining in the pathology domain, self supervised learning offers potential to obtain pre-trained model that preserves the useful information about data in itself. Although one recent study \cite{gildenblat2019self} explored self-supervised similarity learning for pathology image retrieval, much of the self-supervision literature is focused on computer vision applications. A key challenge in applying self-supervision to domain-specific applications is to define the pretext task that will be most beneficial. As such, systematic analysis and derivation of pretext tasks customized for a range histopathology applications would be desirable.

\section{Problem Formulation}
We now define the problem from the perspectives of semi-supervised learning and domain adaptation for pathology image classification. We denote input images as $\mathbf{x}$ and the class labels as $y$.

\paragraph{\textbf{Semi-supervised Learning}}
We consider a set of limited labeled images ${S_L} = \{ ({\mathbf{x}_i},{y_i})\} _{i = 1}^N$, and a set of unlabeled images ${S_U} = \{ ({\mathbf{x}_i})\} _{i = 1}^M$. The semi-supervised framework seeks to leverage unlabeled data ${S_U}$ to enhance the generalizability of learning with labeled data ${S_L}$. Generally, in the semi-supervised setting, both ${S_L}$ and ${S_U}$ come from the same distribution.
\paragraph{\textbf{Domain Adaptation}}
We define a source domain $S$ comprising a set of $n_s $ samples with labeled data $D_s=\{\mathbf{x}_i^{s},y_i^{s}\}_{i=1}^{n_s}$. Likewise, we have a target domain $T$ comprising a set of $n_t$ samples with unlabeled data $D_t=\{\mathbf{x}_i^{t}\}_{i=1}^{n_t}$. The label space of source domain is the same as target domain $Y_s\approx Y_t$. Further, the source and target domains are related but their distributions are distinct.  

\section{Methods}
\begin{figure*}[!t]
\centerline{\includegraphics[width=0.8\textwidth]{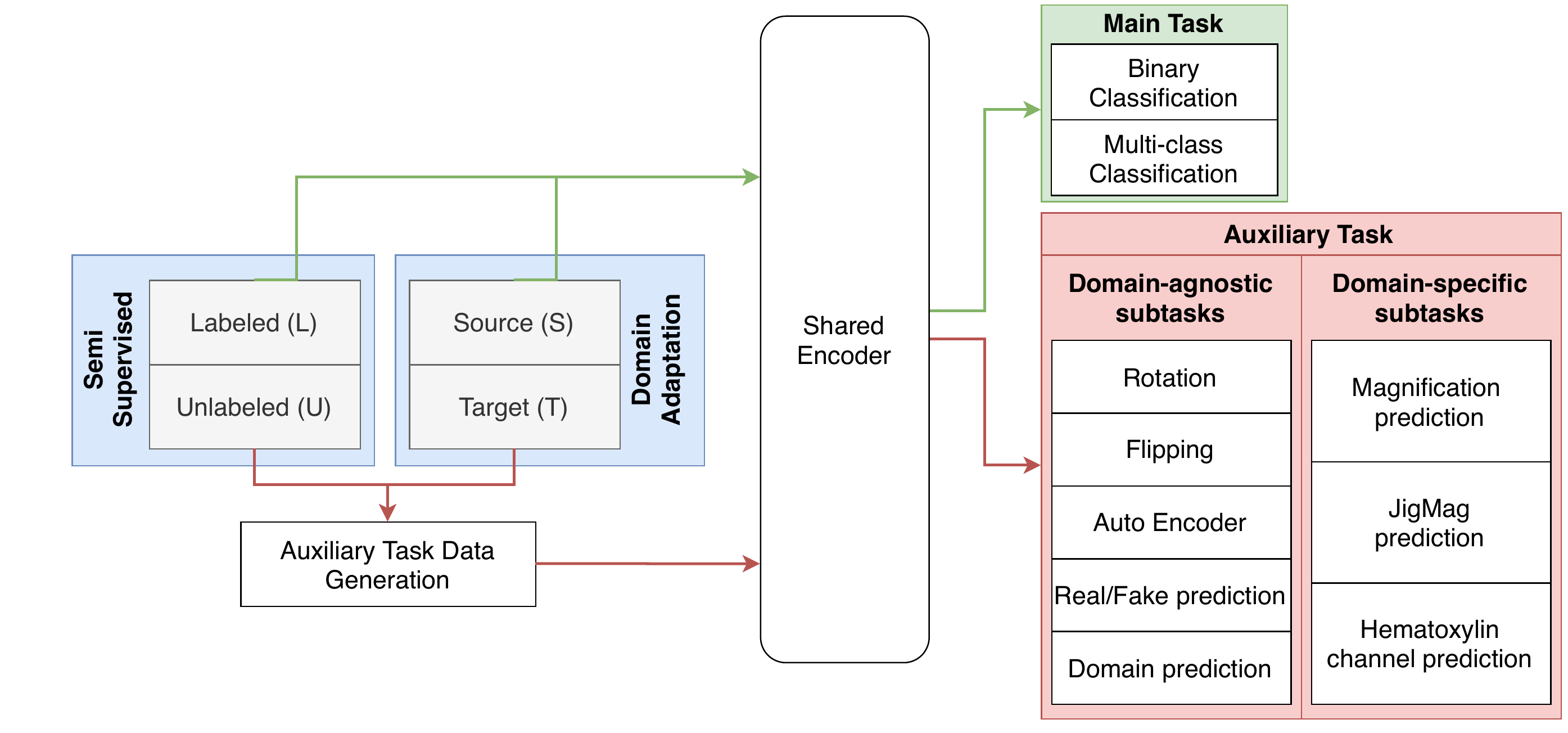}}
\caption{Overview of Self-Path : The framework employs self-supervised auxiliary tasks. Auxiliary tasks can be added to shared encoder to learn useful representations to enhance semi-supervised learning or domain-adaptation.}
\label{fig:overview}
\end{figure*}

Our proposed Self-Path framework is depicted in \cref{fig:overview}. Self-Path follows a two-pronged approach to deal with label scarcity and domain shift by utilizing the power of self-supervision. Specifically, Self-Path employs a multi-task learning approach to learn class discriminative and domain invariant features that would generalize with limited annotated data. The framework can leverage a range of domain specific or domain agnostic pretext tasks, is amenable to adversarial or non-adversarial training, and allows flexibility to incorporate components from semi-supervised, generative learning and domain adaptation approaches. In the following sections, we formally lay out the multi-task learning objective and detail the pretext tasks that are used for along with the main task.  

\subsection{Multi-task learning}
Our proposed approach trains the model using different learning tasks in conjunction. This helps with model generalization and also improves the performance with respect to our primary (main) task. The auxiliary/secondary/side tasks plug into a shared encoder that learns common features along with the main task. Each task usually has a separate head connected to the common encoder and all tasks are optimized simultaneously. Formally:
\begin{equation}
\begin{array}{*{20}{l}}
{\mathop {{\rm{argmin}}}\limits_{{\theta _c},{\theta _e},\;{\theta _{{p_1}}},..{\theta _{{p_{_k}}}}} \frac{1}{{{n^l}}}\sum\limits_i^{{n^l}} {{L_c}} ({H_c}({H_e}({\bf{x}}_i^l)),\;{y_i})\;}\\
{ + \frac{1}{{{n^l}}}\sum\limits_{k = 1}^K {{\alpha _{{p_k}}}\sum\limits_i^{{n^l}} {{L_{{p_k}}}} ({H_{{p_k}}}({H_e}(\widetilde {\bf{x}}_i^l),\;r_{ik}^l)} \;\;}\\
{ + \frac{1}{{{n^u}}}\sum\limits_{k = 1}^K {{\alpha _{{p_k}}}\sum\limits_i^{{n^u}} {{L_{{p_k}}}} ({H_{{p_k}}}({H_e}(\widetilde {\bf{x}}_i^u),\;r_{ik}^u)} }
\end{array}
\end{equation}
where $r$ is the label for pretext task; ${L_c}$ and ${{L_{{p_k}}}}$ are the losses for the main and pretext tasks; $H_e $, is the shared encoder, $H_{c} $ is the function for main task and $H_{p_k}$ is the function of $k^{th}$ pretext task; $\theta_c $, $\theta_e $  and $\theta_{pn}$ are parameters of main task classifier, shared encoder and pretext tasks, respectively; ${\alpha _{{p_k}}}$ indicates weights for different tasks; and $n^{l}$ and $n^{u}$ indicate numbers of labeled and unlabeled samples. When this model is used for semi-supervised learning, the labeled and unlabeled data come from the same domain. When used for domain adaptation, the labeled data comes from source domain and unlabeled data comes from target domain.

\subsection{Self-Supervision}
The self-supervision components entail a range of pretext tasks that can leverage the information from unlabeled data to improve performance in the main task that has few labels for supervision. Our network setup utilizes various both domain specific and domain agnostic self-supervised tasks. 

Consider a transformation function $g(x,r)$ that can be applied on the input $\mathbf{x}$ to transform it to ${\bf{\tilde x}}$ where ${\bf{\tilde x}}$ is served as the input of network and $r$ is the label for that input. $r$ is a predefined value that function $g(.)$ takes alongside with $x$ to generate desired output. For example, if we assume that $g(.) $ is a rotation function, $\mathbf{x}$ is the input image and r is rotation values ($r\in\lbrack1,3\rbrack\; $, rotates images by $r\times90^\circ $), then the output is rotated image, where the objective function for self-supervised classification task is:
\begin{equation}
\label{self_supervised loss}
L_{self}=-E_{x\sim D_{real}}\lbrack\log p_S(r\vert g(x,r))\rbrack
\end{equation}

 \subsection{Histology domain specific auxiliary tasks for self-supervision}
The main factors that differentiate between tissue types or disease conditions in a histopathology image are the shape, morphology and arrangement of nuclei. Therefore, to enable the network to learn semantic representations related to shape of objects, context, location and more precisely shape of nuclei from both target (unlabeled) and source (labeled) domains, we introduce three histology-specific pretext tasks, as below.

\begin{figure*}[!t]
\centerline{\includegraphics[width=0.7\textwidth]{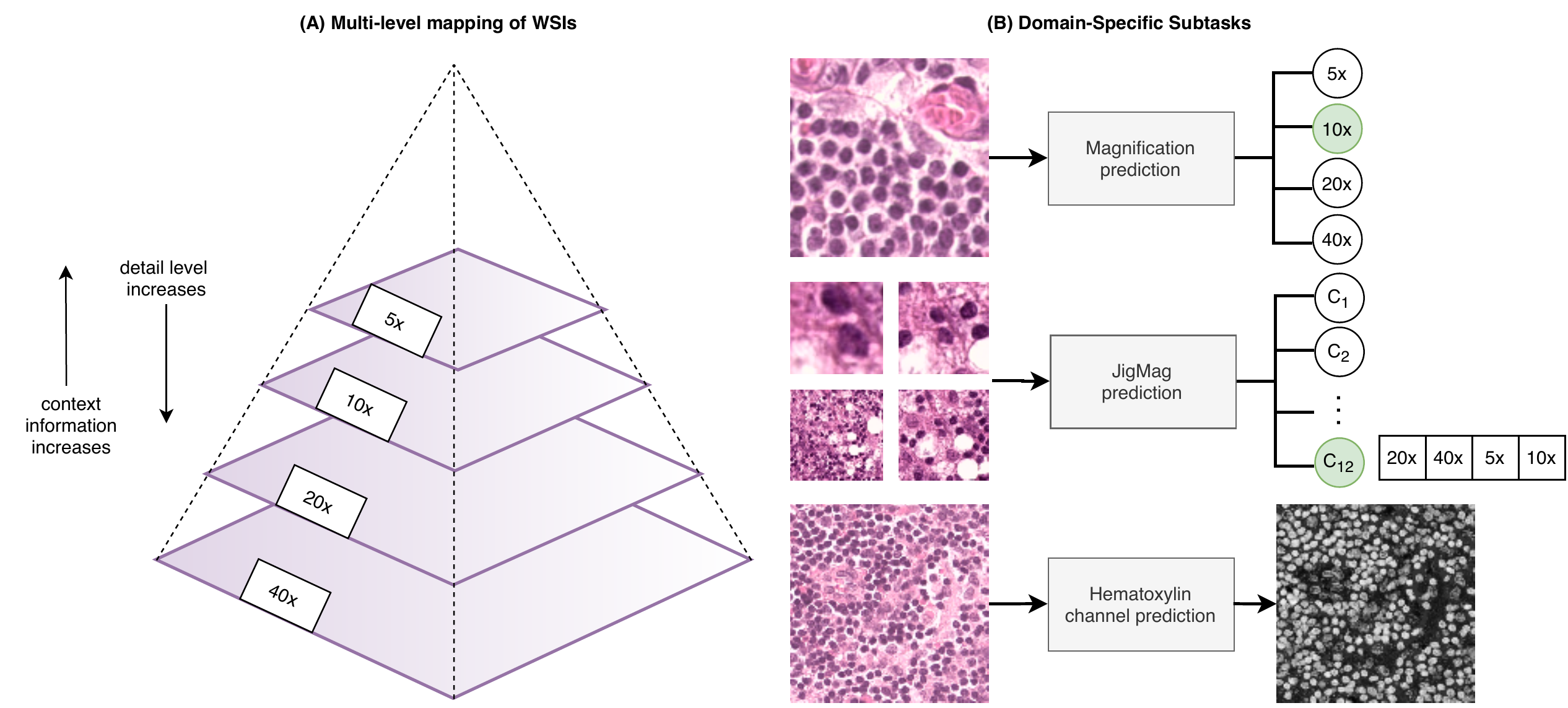}}
\caption{(A) Whole slide images in pathology slides organized hierarchically - each level trades-off between level of detail and context information (B) Domain specific sub-tasks created for Self-Path.}
\label{fig:domain_specific}
\end{figure*}

\subsubsection{Magnification prediction}
Histopathology images are often generated and viewed at various standard magnification levels. Considering the fixed image size, higher magnifications provide more details but less context, whereas lower magnification allows less details but more context of tissue region. By looking at the size and their extent in the images, a specialist can guess the magnification level of images \footnote{Magnification levels and their corresponding resolutions vary for each scanner. However by observing one particular magnification of an image, the other magnification can be perceived easily for the same scanner}. Pathologists assessing an image tend to look at different magnification levels (by zooming out and in on WSIs) to obtain useful information. Therefore, the core idea is to learn semantic information by asking the model to guess it's magnification level. To successfully predict this, the network would have to learn the content information within the image.  We define a pretext task $H_{mag}$ for classifying input to 4 magnification levels (40$\times$, 20$\times$, 10$\times$ and 5$\times$). Images are extracted from these magnification levels \cref{fig:domain_specific}(A), and then fed to the network. If a magnification level is not available, it is obtained by (bilinear) resizing the image from other magnification levels.

\subsubsection{Solving Magnification Puzzle (jigmag)}
The task of retrieving the original image from its shuffled parts is a basic problem of pattern recognition that is commonly identified with jigsaw \cite{freeman1964apictorial}. In the deep learning domain, \cite{noroozi2016unsupervised} proposed to use CNN for solving jigsaw puzzle, where set of jigsaw puzzle permutation is defined and an index is assigned to each of them. 

For solving jigsaw puzzle, the network should concentrate on the differences between tiles and their positions while avoiding to learn low level statistics \cite{noroozi2016unsupervised}. In this way, the network will learn global semantic representation of images.
In histology, objects are smaller compared to the natural images, and there is not specific ordering among each objects (for example, the relative positions of different parts of dog in the natural images is consistent, however we do not have a similar concept in histology), therefore, solving jigsaw puzzle is really challenging and it would not be beneficial for learning useful representations. 
We propose to create a puzzle to reflect magnification and the context of histology images. Being aware of object sizes and the contextual information they provide, can be helpful in the procedure of decision making. Hence, we propose a task for solving magnification puzzle, where image consists of image tiles with various magnification and the network task is to predict their arrangement.
We use this task in conjunction with the main task to learn representations which are reflective of the whole data distribution and enhance the model performance for the main task.

We define $v$ as a vector of image orders in a 2$\times $2 grids where each grid includes a specific magnification. For example $v\;=\;\lbrack0,\;1,\;2,\;3\rbrack $ defines that image with magnification 5$\times$ is on top left corner, 10$\times$ is on top right and so on. As objects with various shapes are present in the pathology, by supervising network in this manner, we force the network to learn representation meaningful for improving the performance for domain adaptation or semi-supervised learning. For this task, 12 different orders of magnification are considered. The loss function for classification tasks is cross entropy and for pixel-wise tasks L1 loss is used. 

\subsubsection{Hematoxylin Channel Prediction}
Nuclei play an important role in H\&E images for deciding on the cancer type and basically are the fundamental part of tissue. Pathologist diagnosis is based on considering nuclei and their arrangements in the image. Therefore, we can learn some useful representations by making the network aware of the nuclei and their morphological information in the image. To do so, the network can be supervised by the nuclear segmentation. However, for most dataset annotating nuclei needs extra time and labour, which makes it infeasible. One way to obtain the rough segmentation of nuclei to extract hematoxylin channel from H\&E images.
In H\&E images, hematoxylin turns the palish color of nuclei to blue and eosin changes the color of other contents to pink. Color deconvolution methods have been applied to specifically identify cell nuclei in H\&E images. Therefore by extracting hematoxylin channel, one can locate the nuclei and their approximate shape. To this end, approach in \cite{ruifrok2001quantification} is used to extract hematoxylin channel as a target for the self-supervision task $H_{hem}$. Usually, since the nuclei shape and their presence indicate the presence of malignancy, this task helps to learn semantic shared features which are potentially related to nuclei and disease progression and can be served for both semi-supervised and domain adaptation applications. The values of hematoxylin channel are scaled in range [0,1] and mean absolute loss is used for optimizing this task. 

\subsection{Domain agnostic self-supervision tasks}
\label{sec:dom_agnos}
To aid with classification, segmentation and image retrieval tasks, various pretext tasks like rotation prediction, flipping, image reconstruction have been studied in the literature \cite{gidaris2018unsupervised,jing2020self}. These were however, not tailored for pathology data. Here, we systematically study and benchmark efficacy of these pretext tasks for semi-supervised learning and domain adaptation in histopathology applications.

\subsubsection{Predicting image rotation}
For rotation prediction, the input image is rotated with degrees of $0^\circ$, $90^\circ$, $180^\circ$ and $270^\circ$ corresponding to the labels 0, 1, 2 and 3, respectively. 

\subsubsection{Predicting image flipping}
The label assigned to the horizontal flipping of image is 1 and 0 if not flipped. 

\subsubsection{Autoencoder} 
For reconstructing the image, a convolutional decoder is used on top of the feature extractor, similar to one for predicting hematoxylin channel. 

\subsubsection{Real vs Fake Prediction (generative)}

From the given unlabeled samples, a generative network (auxiliary task generation block in \cref{fig:overview}) can be trained in an adversarial fashion to generate samples that look like the original samples. The shared encoder can be used to extract features and in this context, the main task is image classification on the labeled subset and the auxiliary task is real vs fake prediction for a given set of real and generated images. In this setting, we found that it is easier to use a simpler encoder/discriminator similar to the GAN in \cite{salimans2016improved} for convergence. Formally, real images are drawn from distribution $D_{Real}$, and the generator function learns the distribution $D_{gen}$ where the goal is to align this two distributions ($D_{gen}\sim D_{real} $). The generator $G(.)$ takes predefined noise variables $z$ from a uniform distribution $D_{noise}$. The objective function is defined as:

\begin{equation}
\begin{array}{*{20}{l}}
{{L_{dis}} =  - {{\rm{E}}_{x \sim \;{D_{real}}}}[\log [1 - {H_{Dis}}({H_e}(x))]]}\\\\
{ - {{\rm{E}}_{x \sim \;{D_{gen}}}}[\log [{H_{Dis}}({H_e}(x))]]}\\\\
{{L_{gen}} = ||{{\rm{E}}_{x \sim {D_{real}}}}|{H_e}(x)| - {{\rm{E}}_{z \sim {D_{noise}}}}|{H_e}(G(z))|{|_1}}
\end{array}
\end{equation}

Where $L_{gen}$  and $L_{dis}$  are the generator and discriminator losses, respectively. ${H_e}(x) $ is the feature from intermediate layer of feature extractor (last layer before fully connected layers) and ${H_{Dis}}({H_e}(x))$ is the output of the discriminator (fake/real head).
    
\subsubsection{Domain prediction}
Domain prediction can enable representation learning for domain adaptation. Deep learning approaches employing domain adversarial neural networks (DANN) seek to perform domain adaptation by learning feature representations that are not distinguishable by domain discriminator \cite{ganin2016domain}. DANN includes a minimax game where discriminator $H_d$ (domain prediction head) is trained to distinguish between the source and target domain, and the feature extractor is simultaneously trained to confuse the discriminator. Therefore, to extract  domain-invariant features $f$, the parameters of feature extractor $\theta_e$ (shared encoder in the multi-task setup ) are learned by maximizing the loss of domain discriminator $L_d$, while parameters of the domain discriminator are learned by minimizing the loss of domain discriminator. Parameters of the main task $H_c$ are also minimized to ensure good performance on the main task. Formally:
\begin{equation}
\begin{array}{l}\underset{\theta_c,\theta_{f\;\;\;\;\;\;}\theta_d}{\text{argminmax}}\frac1{n^{s}}\sum_{i=0}^{n^{s}}L_c(H_c(H_e(\mathbf{x}_i^{s})),\;y_i)\;+\\\;\;\;-\frac{\alpha_d}{n^{s}+n^{t}}(\sum_{i=1}^{n^{s}+n^{t}}L_d(H_d(H_e(\mathbf{x}_i)),\;d_i)\end{array} 
\end{equation}
Where $d_i$ is the domain label for $\mathbf{x}_i $ and $\alpha_d$ is a coefficient for discriminator loss. The domain confusion practically is applied using Gradient Reversal Layer (GRL), where the gradients of $L_d $ with respect to gradients of feature extractor parameters ${\theta _e}$ ($\frac{{\partial {L_d}}}{{\partial {\theta _e}}}$) are reversed during back-propagation.

\section{Experiments}

\subsection{Datasets}
\subsubsection{Camelyon16}
We used the Camelyon 16 challenge dataset that comprises 399 H\&E stained histological images of lymph nodes in the breast. This dataset contains images with the breast cancer metastasis in the lymph nodes and WSIs were acquired from 2 different centers of Radboud University Medical Center (RUMC) and University Medical Center Utrecht (UMCU). RUMC images were generated by a digital slide scanner (Pannoramic 250 Flash ; 3DHISTECH) with a 20$\times$ objective lens (0.243 $\mu m$ $\times$ 0.243 $\mu m$) and UMCU images were produced using a digital slide scanner (NanoZoomer-XR Digital slide scanner C12000-01; Hamamatsu Photonics) with a 40$\times$ objective lens (0.226$\mu m$ $\times$ 0.226 $\mu m$). This dataset is split into 270 training and 129 test images. 34 WSIs of training set are used for validation. The tumor regions are exhaustively annotated by pathologists. We randomly extracted patches from normal and tumor regions for our experiments.  

\subsubsection{LNM-OSCC}
LNM-OSCC is an in-house dataset comprising images of Oral Squamous Cell Carcinoma metastasized to the cervical lymph nodes. The H\&E WSIs for this dataset were acquired from two hospitals using two different scanners. 98 WSIs were scanned with 40$\times$ objective lens using IntelliSite Ultra Fast Scanner (0.25 $\mu m$/pixel) at University Hospital Conventry and Warwickshire (UHCW). 119 of WSI were produced at the School of Medical Dentistry in Sheffield University by  Aperio/Leica CS2 with 20$\times$ objective lens ( 0.2467 $\mu m$/pixel). From these 217 samples, 100 WSIs are used for training purposes and 14 for validation and 103 for test. For training and validation set, tumor regions were not exhaustively annotated.

\subsubsection{Kather}
This dataset contains 100K image patches from HE stained histological WSIs of human colorectal cancer (CRC) and normal tissue. For this dataset only patches were available without accessing to WSIs. This dataset covers 9 tissue classes: Adipose (ADI), background (BACK), debris (DEB), lymphocytes (LYM), mucus (MUC), smooth muscle (MUS), normal colon mucosa (NORM), cancer-associated stroma (STR), colorectal adenocarcinoma epithelium (TUM). Images in this dataset are 224 $\times$ 224 where we resize them to 128 $\times$ 128 for our experiments. 20\% of this dataset is considered as validation set and another set of  7180 image patches is used for test.

\cref{table:data stats} shows the number of patches extracted for Camleyon16 and LNM-OSCC dataset. For our main task patches of 128 $\times$ 128 at 10$\times$ are extracted from LNM-OSCC and Camelyon16.

\begin{table}
\centering
\caption{Number of WSIs and patches in each dataset.}
\begin{tabular}{l|lccc}
\multicolumn{1}{l}{}        &         & \multicolumn{1}{l}{Train} & \multicolumn{1}{l}{Validation} & \multicolumn{1}{l}{Test}  \\ 
\hhline{~~---}
\multirow{2}{*}{Camelyon16} & WSIs    & 236                       & 34                             & 129                       \\
                            & patches & 67054                     & 15586                          & 16562                     \\ 
\hline
\multirow{2}{*}{LNM-OSCC}   & WSIs    & 100                       & 14                             & 103                       \\
                            & patches & 55416                     & 7224                           & 14472                     \\ 
\hline
Kather                      & patches & 79994                     & 20006                          & 7180                     
\end{tabular}
\label{table:data stats}
\end{table}

\begin{figure*}[!t]
\centerline{\includegraphics[width=.6\textwidth]{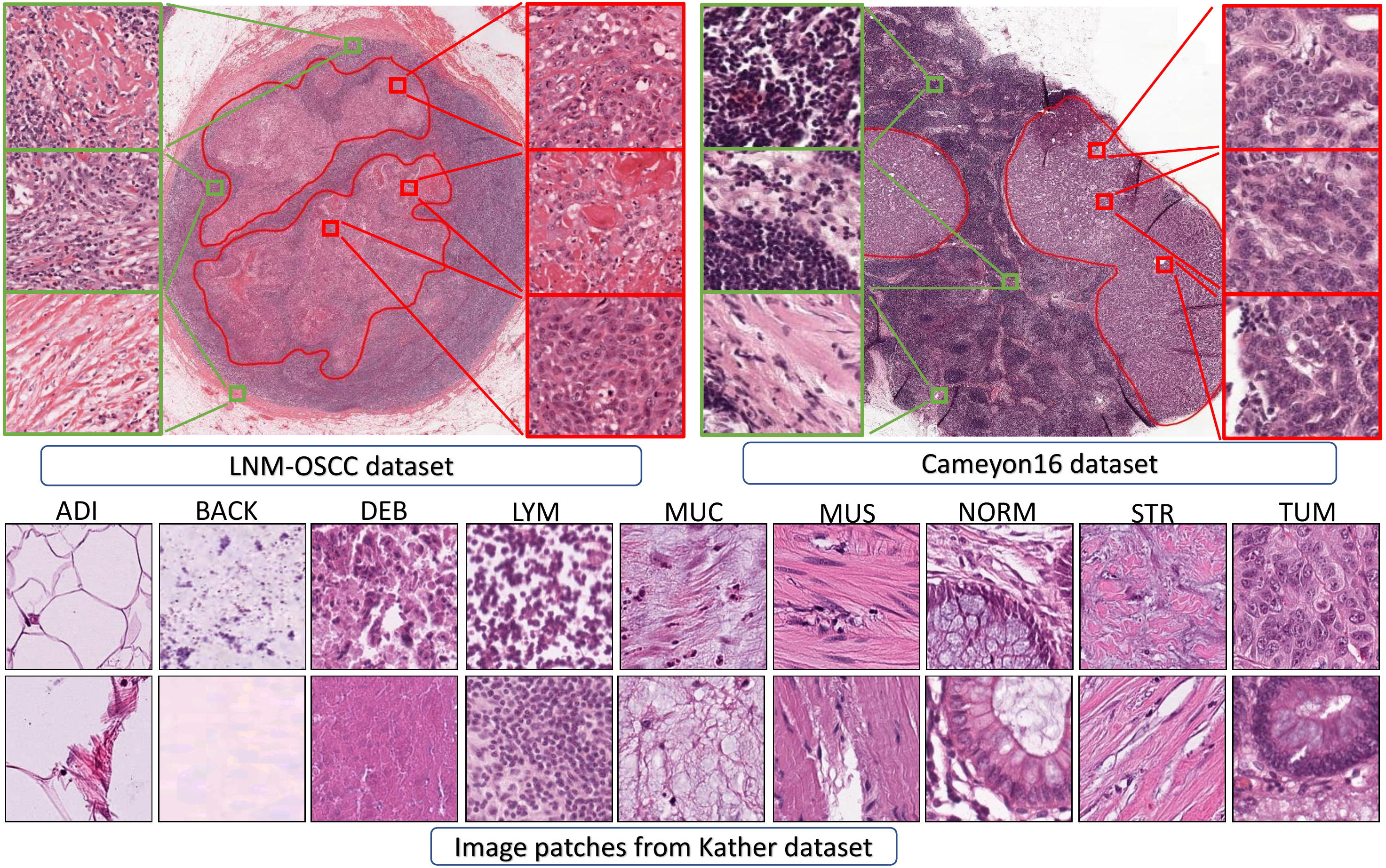}}
\caption{Exemplar images of different datasets that are used in this study. Red and green boxes denote the tumor and normal image patches.}
\label{fig1}
\end{figure*}

\subsection{Experimental setup}
\subsubsection{Networks}
We chose Resnet50 \cite{he2016deep} as the feature extraction backbone for our experiments. The classifier head consists of adaptive average pooling which is followed by fully connected layer and softmax. The decoder head for reconstructing image and predicting hematoxylin channel is similar to the UNet decoder \cite{ronneberger2015u} without using any skip connections. While using the real vs fake auxiliary task for image generation, we utilize the architecture presented in \cite{salimans2016improved} and find that this simpler feature extractor allows easy and robust convergence for the image generator.

\subsubsection{Implementation details}
When Resnet50 is used as the shared encoder, the network was trained for 200 epochs. The batch size for our experiments was 64, and Adam optimizer was used. The learning rate was set at ${10^{ - 3}}$. An epoch is defined as one full step through all unlabeled data. For experiments related to fake/real prediction, number of epoch and batch size were 500 and 32 respectively. Adam optimizer with learning rate of $3 \times {10^{ - 4}}$ was utilized. The weighting coefficient for all task was set to 1 for all experiments.

\subsection{Semi-Supervised Experiments}
Here, we compare the  effect of different self-supervision tasks for semi-supervised learning. We compare our models against the popular semi-supervised benchmarks, namely Mean Teacher \cite{tarvainen2017mean} and VAT \cite{miyato2018virtual}. 
We also compare with teacher-student chain \cite{shaw2020teacher} (TSchain), a recent semi-supervised model that was proposed within histology domain, that predicts the pseudo-labels for the unlabeled data and then uses all images for iteratively retraining the model. 
For performance evaluations, we follow the typical protocol of varying the annotation budget for the training set while maintaining a fixed validation set, and reporting AUCs on the test set. For our experiments, we employed Resnet50 as the backbone for all models, used 3 iterations and report the average results from the 3 runs with different random seeds.

\subsubsection{Comparisons of semi-supervised learning on LNM-OSCC dataset}
\begin{table*}
\centering
\caption{Performance of different models for different percentages of annotation budgets for LNM-OSCC dataset. The supervised upper bound performance when using all labeled data is 98.4\%.  Numbers in the parenthesis indicates number of patches used for each budget of annotation.}
\label{table: oscc-result}
\begin{tabular}{lc|c|c|c|c} 
\hline
Labeled WSIs          & \multicolumn{1}{c}{1\%(134)}                 & \multicolumn{1}{c}{2\%(1024)}                & \multicolumn{1}{c}{5\%(1880)}                & \multicolumn{1}{c}{10\%(3334)}               & 20\%(7558)                 \\ 
\cline{2-6}
                      & AUC-ROC(\%)                                  & AUC-ROC(\%)                                  & AUC-ROC(\%)                                  & AUC-ROC(\%)                                  & AUC-ROC(\%)                \\ 
\cline{2-6}
supervised baseline   & 73.4 $\pm$ 2.0                              & 76.1 $\pm$ 5.3                              & 85.3 $\pm$ 6.3                              & 86.3 $\pm$ 2.7                              & 96.3 $\pm$ 0.3            \\
mean teacher \cite{tarvainen2017mean}          & 75.1 $\pm$ 4.5                              & 78.4 $\pm$ 5.6                              & 86.2 $\pm$ 7.6                              & 91.4 $\pm$ 1.2                              & \textbf{97.4 $\pm$ 0.3}   \\
VAT \cite{miyato2018virtual}              & 74.5 $\pm$ 5.6                              & 77.4 $\pm$ 3.3                              & 85.3 $\pm$ 4.3                              & 92.1 $\pm$ 1.2                              & 96.5 $\pm$ 0.9            \\
TS chain \cite{shaw2020teacher}             & 75.3~$\pm$~2.4                              & 79.3~$\pm$~2.5                              & \multicolumn{1}{c}{85.2~$\pm$~3.1}          & 94.1~$\pm$~1.7                              & 97.2~$\pm$~0.2            \\ 
\cline{2-6}
\multicolumn{1}{c}{}  & \multicolumn{5}{c}{Domain agnostic self-supervised tasks}                                                                                                                                                         \\ 
\cline{2-6}
rotaion               & 74.5 $\pm$ 5.6                              & 76.3 $\pm$ 4.2                              & 88.4 $\pm$ 1.5                              & 93.2 $\pm$ 0.3                              & 96.2 $\pm$ 0.1            \\
flipping              & 74.6 $\pm$ 4.0                              & 74.2 $\pm$ 5.3                              & 85.3 $\pm$ 4.1                              & 91.4 $\pm$ 0.4                              & 94.2 $\pm$ 0.4            \\
autoencoder           & 73.0 $\pm$ 6.5                              & 75.1 $\pm$ 3.5                              & 84.2 $\pm$ 3.3                              & 90.3 $\pm$ 1.5                              & 94.3 $\pm$ 0.2            \\ 

generative            & 73.4 $\pm$ 7.1                              & 79.3 $\pm$ 4.1                              & \textbf{90.3 $\pm$ 2.4}                     & \textbf{95.4 $\pm$ 0.2}                     & 97.1 $\pm$ 0.3            \\
\cline{2-6}
                      & \multicolumn{5}{c}{Domain specific self-supervised tasks}                                                                                                                                                         \\ 
\cline{2-6}
magnification         & 76.3 $\pm$ 4.0                              & 76.6 $\pm$ 3.6                              & 87.4 $\pm$ 2.3                              & 92.5 $\pm$ 0.2                              & 94.1 $\pm$ 0.4            \\
jigmag                & \textbf{80.1 $\pm$ 5.1}                     & \textbf{82.1 $\pm$ 5.2}                     & 89.5 $\pm$ 4.6                              & 92.2 $\pm$ 0.4                              & 96.3 $\pm$ 0.3            \\
hematoxylin           & 75.3 $\pm$ 7.6                              & 80.2 $\pm$ 5.3                              & 87.5 $\pm$ 1.2                              & 94.4 $\pm$ 1.3                              & \textbf{97.4 $\pm$ 0.5}   \\ 
\hline
Best self-supervised~ & \multicolumn{1}{c}{\textbf{80.1~$\pm$~5.1}} & \multicolumn{1}{c}{\textbf{82.1~$\pm$~5.1}} & \multicolumn{1}{c}{\textbf{90.3~$\pm$~2.4}} & \multicolumn{1}{c}{\textbf{95.4~$\pm$~0.2}} & \textbf{97.4~$\pm$~0.5}  
\end{tabular}
\label{table:oscc-result}
\end{table*}

Performance of each of the self-supervised tasks on LNM-OSCC dataset are reported in \cref{table:oscc-result}. we have evaluated the model performance in terms of AUC-ROC (Area Under the Receiver Operating Characteristics) for different budget of annotations. We have used 1, 4, 5, 10 and 20 labeled WSIs out of 100 training WSIs and the remaining images are used as unlabeled images in the semi-supervised models. Supervised baseline is only trained on labeled images without utilizing any unlabeled data. 

We observe from  \cref{table:oscc-result} that in case of having very few annotations, domain specific self-supervised tasks have better performance compared to supervised baseline, mean teacher,  VAT and TS chain. For instance, when we are dealing with only 1\% (1 labeled WSI) and 4\% (4 labeled WSIs) of budget of annotations (134 and 1120 labeled patches), jigmag task has the best performance.  Hematoxylin and magnification tasks also outperform domain agnostic tasks and generative tasks at 1\% and 2\% of budget. We can also observe the high performance of generative task (AUC of 95.4\%) when the amount of annotation increases to 5\%. It means that the generated images can help the classifier to boost the performance. Overall,  we observe form the experiments on LNM-OSCC dataset that simple pretext tasks are helpful for enhancing the model performance in the case of limited annotations, with jigmag outperforming other approaches.

\subsubsection{Comparisons of semi-supervised learning on Camelyon16 dataset} 
\begin{table*}
\centering
\caption{Performance of different models for different percentages of annotation budgets for Camelyon16 dataset, the supervised upper bound performance is 94.2\%. Numbers in the parenthesis indicate number of patches used for each budget of annotation. }
\label{table: Camelyon result}
\begin{tabular}{lc|c|c|c|c} 
\hline
Labeled WSIs         & \multicolumn{1}{c}{1\%(600)} & \multicolumn{1}{c}{2\%(1000)} & \multicolumn{1}{c}{5\%(2600)} & \multicolumn{1}{c}{10\%(6400)} & 20\%(13540)               \\ 
\cline{2-6}
                     & AUC-ROC(\%)                  & AUC-ROC(\%)                   & AUC-ROC(\%)                   & AUC-ROC(\%)                    & AUC-ROC(\%)               \\ 
\cline{2-6}
supervised baseline  & 68.3 $\pm$ 5.1               & 74.5 $\pm$ 5.8                & 81.2 $\pm$ 2.5                & 88.4 $\pm$ 2.3                 & 92.1 $\pm$ 0.5            \\
Mean Teacher  \cite{tarvainen2017mean}         & 73.7 $\pm$ 3.8               & 78.5 $\pm$ 2.6                & 84.5 $\pm$ 2.4                & 92.7 $\pm$ 1.9                 & 93.1 $\pm$ 0.9            \\
VAT \cite{miyato2018virtual}                 & 70.9 $\pm$ 5.8               & 77.4 $\pm$ 3.3                & 81.3 $\pm$ 5.2                & 90.3 $\pm$ 2.3                 & 92.8 $\pm$ 1.5            \\
TS chain \cite{shaw2020teacher}            & 74.9 $\pm$ 6.9                & 76.9 $\pm$ 3.2                 & 83.8 $\pm$ 2.1                 & 93.1~$\pm$2.5                  & \textbf{93.9 $\pm$ 1.3}    \\ 
\cline{2-6}
                     & \multicolumn{5}{c}{Domain agnostic self-supervised tasks}                                                                                                 \\ 
\cline{2-6}
rotation             & 69.8 $\pm$ 4.8               & 74.5 $\pm$ 3.1                & 80.4 $\pm$ 2.5                & 90.1 $\pm$ 2.0                 & 92.4 $\pm$ 2.5            \\
flipping             & 70.2 $\pm$ 6.2               & 75.4 $\pm$ 3.5                & 81.6 $\pm$ 5.1                & 89.4 $\pm$ 0.6                 & 92.3 $\pm$ 1.6            \\
autoencoder          & 70.1 $\pm$ 2.4               & 75.6 $\pm$ 4.1                & 82.3 $\pm$ 4.5                & 90.5 $\pm$ 2.3                 & 92.4 $\pm$ 1.1            \\
generative           & 72.5 $\pm$ 5.5               & 77.6 $\pm$ 5.4                & 82.4 $\pm$ 7.2                & 92.6 $\pm$ 3.2                 & 93.6 $\pm$ 1.5            \\ 
\cline{2-6}
                     & \multicolumn{5}{c}{Domain specific self-supervised tasks}                                                                                                 \\ 
\cline{2-6}
magnification        & 77.5 $\pm$ 3.1               & 84.6 $\pm$ 5.2                & 85.1 $\pm$ 3.6                & \textbf{93.2 $\pm$ 3.4}        & 93.4 $\pm$ 2.5            \\
jigmag               & \textbf{81.6 $\pm$ 3.5}      & \textbf{85.4 $\pm$ 2.4}       & \textbf{87.2 $\pm$ 4.3}       & 90.4 $\pm$ 3.1                 & 92.2 $\pm$ 1.5            \\
hematoxylin          & 72.8 $\pm$ 4.6               & 78.3 $\pm$ 4.5                & 84.6 $\pm$ 3.4                & 92.3 $\pm$ 4.1                 & \textbf{93.7 $\pm$ 2.5}            \\ 
\hline
Best Self-supervised & \textbf{81.6 $\pm$ 3.5}      & \textbf{85.4 $\pm$ 2.4}       & \textbf{87.2 $\pm$ 4.3}       & \textbf{93.2 $\pm$ 3.4}        & \textbf{93.7 $\pm$ 2.5}  
\end{tabular}
\label{table:cam-result}
\end{table*}

For this dataset, we have used 2, 4, 8, 20 and 40 labeled WSI out of 236 training WSIs and the performance is reported on the patches extracted from test (129) set. Performance of different methods are reported in the \cref{table:cam-result}. As one can see, similar to LNM-OSCC dataset, domain specific tasks outperform other semi supervised methods. Particularly, jigmag task could improve the performance by 13.3\%, 10.9\% and 6\% at 1\% (2 WSIs), 2\% (4 WSIs) and 5\%  (8 WSIs), respectively.

At 1\% of annotation budget, only magnification and jigmag outperform mean teacher and supervised baseline. Unlike LNM-OSCC that generative model could achieve highest AUC-ROC for some level of annotation budget, here it cannot achieve best performance, but it's performance is competitive with mean teacher and VAT. Similar to LNM-OSCC jigmag could achieve highest performance overall, and the main boost is obtained when we are dealing with very few amount of labeled data.

\subsubsection{Comparisons of semi-supervised learning on Kather dataset} 
\begin{table}
\centering
\caption{Performance of different models for different budget of annotations for kather dataset, the supervised upper bound performance when using all labeled data is 99.4\%.}
\begin{tabular}{lc|c} 
\hline
labeled patches      & \multicolumn{1}{c}{0.1\%(100)} & 1\%(800)                  \\ 
\cline{2-3}
                     & AUC-ROC(\%)                    & AUC-ROC(\%)               \\ 
\cline{2-3}
supervised baseline  & 87.5 $\pm$ 2.0                 & 92.5 $\pm$ 1.2            \\
mean teacher  \cite{tarvainen2017mean}         & 89.1 $\pm$ 1.5                 & 93.9 $\pm$ 0.3            \\
VAT \cite{miyato2018virtual}                 & 88.5 $\pm$ 1.4                 & 92.6 $\pm$ 0.4            \\
TS chain \cite{shaw2020teacher}            & 88.9 $\pm$ 0.3                 & 93.5 $\pm$ 0.2            \\ 
\hline
                     & \multicolumn{2}{c}{Self-supervised tasks}                  \\ 
\hline
generative           & 88.4 $\pm$ 3.5                 & 92.3 $\pm$ 2.6            \\
rotation             & 87.4 $\pm$ 1.6                 & 93.3 $\pm$ 0.4            \\
flipping             & 88.6 $\pm$ 0.8                 & 93.0 $\pm$ 0.9            \\
autoencoder          & 89.3 $\pm$ 1.3                 & 94.3 $\pm$ 1.2            \\
hematoxylin          & 90.3 $\pm$ 0.7                 & 95.1 $\pm$ 0.5            \\ 
\hline
Best self-supervised & \textbf{90.3 $\pm$ 0.7}        & \textbf{95.1 $\pm$ 0.5}   \\
                     & \multicolumn{1}{l}{}           & \multicolumn{1}{l}{}     
\end{tabular}
\label{table:katherresult}
\end{table}
The results of semi-supervised experiments on Kather dataset are reported in \cref{table:katherresult}. Since there are 9 classes in the Kather dataset, Macro AUC-ROC is used for evaluation of classification performance. Unlike the other 2 datasets, only patches were available for this dataset, therefore the annotation budget is considered based on only amount of patches.

In our experiments, we observed when the number of labeled images decreases down to 2\% of whole dataset, the performance of supervised baseline is still high (Macro AUC of 98\%) . Hence using semi supervised approached for that budget of annotations would not add any benefit, so the Macro AUC is calculated at 1\% (800 labeled images) and 0.1\%(100 labeled) of whole annotation budget, where some degradation of Macro AUC can be observed for supervised model.
Moreover, as we didn't have access to WSIs for this dataset to extract large patches or patches at different magnifications, jigmag and magnification self-supervised task could not be carried out on this dataset. 
It is inferred from table \ref{table:katherresult} that at 1\% annotation budget, predicting hematoxylin channel as a self-supervised task improves the performance by 2.8\% and 1.2\% compared to the baseline and mean teacher ,respectively. 

When using 800 labeled data,  using various self-supervised tasks, can again improve performance compared to the baseline. Predicting hematoxylin channel, can also gives the superior performance, which proves the prediction of rough nuclear segmentations can be helpful for semi-supervised learning.


\subsection{Domain adaptation experiments}
For this experiment, we consider  Camelyon16 data set as the labeled set (source domain) and LNM-OSCC dataset as the unlabeled set (target domain). The model is then tested on the test set of LNM-OSCC dataset. We try to solve unsupervised domain adaptation where all labeled data come from the training set of Camelyon16, and LNM-OSCC training data are used without any labels.

Therefore, we want to learn the useful representations that the model can generalize well on the target domain.
In table \cref{table:katherresult} the comparison between different self-supervision tasks and two methods WDGRL \cite{shen2018wasserstein} and DANN \cite{ganin2016domain}, are reported. WDGRL trains a domain critic network to estimate the  Wasserstein distance between the source and target feature representations. The feature extractor network will then be optimized to minimize the estimated Wasserstein distance in an adversarial manner. By iterative adversarial training, we finally learn feature representations invariant to the covariate shift between domains. DANN approach is based on the GRL unit and was mentioned in \cref{sec:dom_agnos}. 

\begin{figure}
    \centering
    \includegraphics[width=.75\columnwidth]{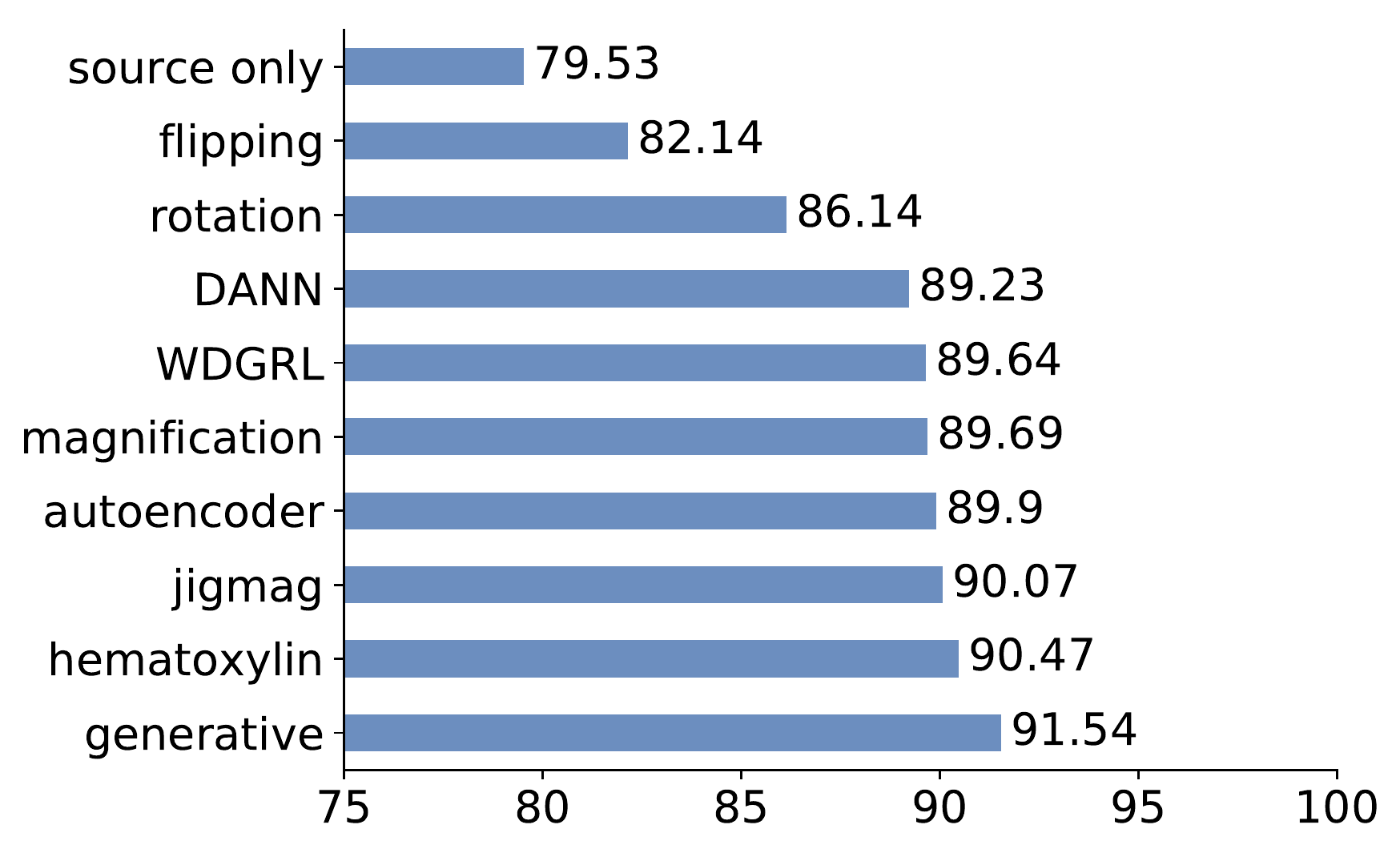}
    \caption{Comparisons of macro AUC-ROC of various methods for  domain adaptation across the datasets Camelyon16 $\rightarrow$ LNM-OSCC}
    \label{fig:barchart}
\end{figure}

Baseline experiment here is the Resnet50 when it is trained by using only Camelyon16 dataset. As shown in \cref{fig:barchart}, predicting domain specific tasks as pretext tasks in the proposed pipeline can help the model to gain the performance by a large margin compared to the baseline. More precisely, predicting magnification, jigmag and hematoxilyn channel, as auxiliary tasks boost performance by 10\% in terms of AUC-ROC. It is also evident that using generative model in our proposed pipeline achieves best performance, which is 2\% higher than WDGRL and 11\% improvement over the baseline  which proves the generated images by generator can contribute to the learning useful domain invariant features. Note that for all experiments that generative model is not included we have used domain prediction with GRL layer. In our experiments, we observed that generative model without using any domain adaptation techniques such as GRL can still achieve very high performance, which might be that generated images and adversarial training compensate for the need of GRL.

\subsubsection{WSI Analysis}
Mainly patch level classification is a prerequisite for WSI-level disease prediction, where during test phase, patches belonging to tissue region are aggregated to construct a heat map for the WSIs. Afterwards, the label for WSI is obtained by post-processing the heat maps.
Here, we evaluate the performance of domain adaptation on the WSIs of LNM-OSCC dataset . To this end, we use our best self-path setting (being generative model) and apply it on the patches extracted from test WSIs. Our method is compared with the baseline being a model trained only on the Camelyon16 dataset.

\begin{figure}[!t]
\centerline{\includegraphics[width=0.8\columnwidth]{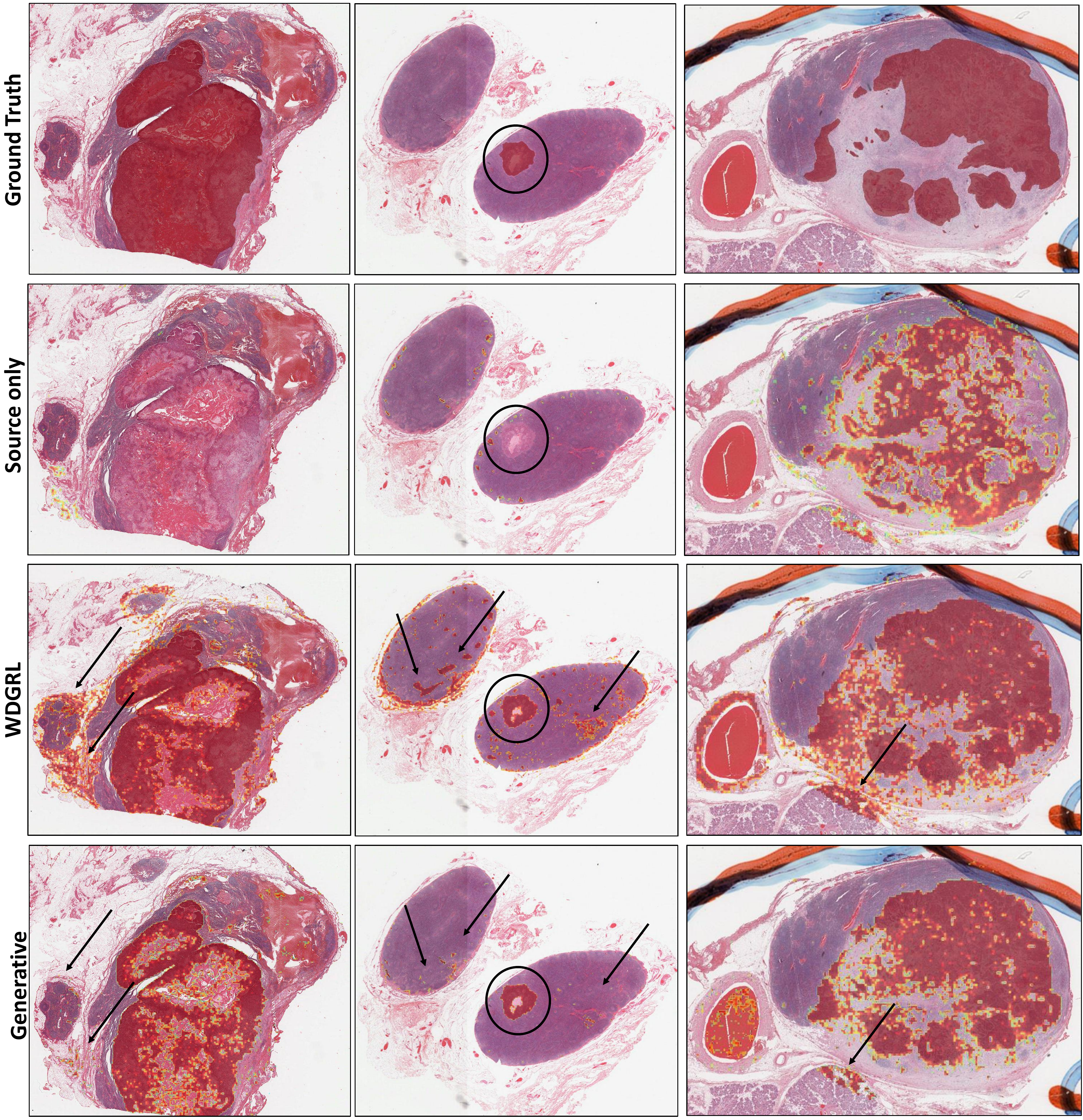}}
\caption{Three WSI samples and their overlaid heatmaps. from top to bottom, first row: the overlaid ground-truth mask, second row: overlaid heat map of model predictions when it is trained using only Camlelyon16 data, third row: Overlaid heatmap of WDGRL predictions and the last row depicts the overlaid predictions of Self-path using generative task. The circle indicates a region which is missed using the baseline model and arrows point to the false positive regions generated by WDGRL where using generative task eliminates those regions.}
\label{fig:heatmap}
\end{figure}

For heat map generation, first, patches of 128 $\times$ 128 at 10$\times$ magnification with overlap of 50\% are extracted from tissue regions of WSIs, second, the  prediction of each patches are aggregated together to build the final heat map of WSIs.
For all models in this section we do the same following post processing: 10 morphological and geometrical feature are extracted from objects within binarized heat map at 3 three thresholds of 0.25, 0.5 and 0.9. Then mean, stddev, minimum and maximum of object features for each WSI are calculated. Therefore, in total 120 features are used for constructing feature vectors. Afterwards, random forest algorithm is used for classification of the features. Finally, the model is evaluated on the test set of LNM-OSCC. 
For these experiments, we have chosen generative model as the best performing domain adaptation approach and compared it against source-only and WDGRL.
The result are shown in \cref{table:wsi_level_DA}. Using only Camelyon16 data results in AUC-ROC of 75.2 whereas utilizing unlabeled images by generative model task can gain the performance by 15.2\% which is a large improvement. This again proves the capability of generated images and adversarial training of GAN for achieving high performance in both patch-level and WSI-level.
2\% improvement of generative model on the patch-level AUC-ROC is translated to 5\% boost in the performance in terms of WSI classification when it is compared to WDGRL. This improvement is also evident in the visual result (\cref{fig:heatmap}).
In \cref{fig:heatmap}, 3 WSIs with their corresponding overlaid heat maps are shown. As one can see in this figure, model that is trained using only source labeled data (middle column) has many false negative and tumor region are completely missed. However, using WDGRL and generative mode  whereas using unlabeled data in our pipeline could increase true positive while decreasing false negatives. The lower performance of WDGRL is mainly due to the lower performance in the patch-level classification task because of large positives which can be clearly seen in the \cref{fig:heatmap}.
\begin{table}
\centering
\caption{Camelyon16 $ \to $ LNM-OSCC domain adaptation results on the WSI-level. The upper bound performance using all labels for target domain in supervised fashion is 93.3\%.}
\label{table:wsi_level_DA}
\begin{tabular}{lcc} 
\hline
            & \multicolumn{1}{l}{AUC-ROC(\%)} & \multicolumn{1}{l}{Average Precision(\%)}  \\ 
\hline
source-only & 75.2                        & 81.7                                   \\ 
WDGRL       & 85.8                        & 91.6                                   \\ 
generative  & 90.4                        & 95.2                                   \\
\cmidrule(lr){1-3}
\end{tabular}
\end{table}

\section{Conclusion}

In this paper, we proposed Self-Path -- a generic framework based on self-supervision tasks for histology image classification -- to address the challenge of limited annotations in the area of computational pathology. We introduced 3 novel self-supervision tasks to cater for the contextual, multi-resolution and semantic features in pathology images. We showed that such domain specific self-supervision tasks can improve the classification performance for both semi-supervised learning and domain adaptation. Moreover, we thoroughly investigated general self-supervised approaches such as generative models within this pipeline and showed that using the domain-specific tasks, despite being simple and easy to implement, can improve the performance in most scenarios when dealing with limited annotation budget or domain shift. In particular, we note that the jigmag self-supervision can be extremely helpful when the amount of labeled data is very small. Unlike baseline methods that are highly dependent on the values of hyperparameters, our method does not require exhaustive tuning of hyperparameters to achieve strong performance. Self-Path can be applied to other problems in computational pathology, where annotation budget is often limited or large amount of unlabeled image data is available. Another future direction could be employing other self-supervision tasks such as predicting the Eosin channel or a combination of Hematoxylin and Eosin after estimating the two channels, rather than keeping them fixed, and increasing the jigmag grids to incorporate wider and complex puzzles for the network to solve. 

\bibliographystyle{ieeetr}

\bibliography{ref}

\begin{thebibliography}{10}

\bibitem{skrede2020deep}
O.-J. Skrede, S.~De~Raedt, A.~Kleppe, T.~S. Hveem, K.~Liest{\o}l, J.~Maddison,
  H.~A. Askautrud, M.~Pradhan, J.~A. Nesheim, F.~Albregtsen, {\em et~al.},
  ``Deep learning for prediction of colorectal cancer outcome: a discovery and
  validation study,'' {\em The Lancet}, vol.~395, no.~10221, pp.~350--360,
  2020.

\bibitem{tarvainen2017mean}
A.~Tarvainen and H.~Valpola, ``Mean teachers are better role models:
  Weight-averaged consistency targets improve semi-supervised deep learning
  results,'' in {\em Advances in neural information processing systems},
  pp.~1195--1204, 2017.

\bibitem{miyato2018virtual}
T.~Miyato, S.-i. Maeda, M.~Koyama, and S.~Ishii, ``Virtual adversarial
  training: a regularization method for supervised and semi-supervised
  learning,'' {\em IEEE transactions on pattern analysis and machine
  intelligence}, vol.~41, no.~8, pp.~1979--1993, 2018.

\bibitem{peikari2018cluster}
M.~a. e.~a. Peikari, ``A cluster-then-label semi-supervised learning approach
  for pathology image classification,'' {\em Scientific reports}, vol.~8,
  no.~1, pp.~1--13, 2018.

\bibitem{li2018based}
J.~Li, W.~Speier, K.~C. Ho, K.~V. Sarma, A.~Gertych, B.~S. Knudsen, and C.~W.
  Arnold, ``An em-based semi-supervised deep learning approach for semantic
  segmentation of histopathological images from radical prostatectomies,'' {\em
  Computerized Medical Imaging and Graphics}, vol.~69, pp.~125--133, 2018.

\bibitem{sparks2016out}
R.~Sparks and A.~Madabhushi, ``Out-of-sample extrapolation utilizing
  semi-supervised manifold learning (ose-ssl): content based image retrieval
  for histopathology images,'' {\em Scientific reports}, vol.~6, p.~27306,
  2016.

\bibitem{noroozi2016unsupervised}
M.~Noroozi and P.~Favaro, ``Unsupervised learning of visual representations by
  solving jigsaw puzzles,'' in {\em European Conference on Computer Vision},
  pp.~69--84, Springer, 2016.

\bibitem{jing2020self}
L.~Jing and Y.~Tian, ``Self-supervised visual feature learning with deep neural
  networks: A survey,'' {\em IEEE Transactions on Pattern Analysis and Machine
  Intelligence}, 2020.

\bibitem{zhai2019s4l}
X.~Zhai, A.~Oliver, A.~Kolesnikov, and L.~Beyer, ``S4l: Self-supervised
  semi-supervised learning,'' in {\em Proceedings of the IEEE international
  conference on computer vision}, pp.~1476--1485, 2019.

\bibitem{graham2020dense}
S.~Graham, D.~Epstein, and N.~Rajpoot, ``Dense steerable filter cnns for
  exploiting rotational symmetry in histology images,'' {\em arXiv preprint
  arXiv:2004.03037}, 2020.

\bibitem{van2020survey}
J.~E. Van~Engelen and H.~H. Hoos, ``A survey on semi-supervised learning,''
  {\em Machine Learning}, vol.~109, no.~2, pp.~373--440, 2020.

\bibitem{lee2013pseudo}
D.-H. Lee, ``Pseudo-label: The simple and efficient semi-supervised learning
  method for deep neural networks,'' in {\em Workshop on challenges in
  representation learning, ICML}, vol.~3, 2013.

\bibitem{berthelot2019mixmatch}
D.~Berthelot, N.~Carlini, I.~Goodfellow, N.~Papernot, A.~Oliver, and C.~A.
  Raffel, ``Mixmatch: A holistic approach to semi-supervised learning,'' in
  {\em Advances in Neural Information Processing Systems}, pp.~5049--5059,
  2019.

\bibitem{laine2016temporal}
S.~Laine and T.~Aila, ``Temporal ensembling for semi-supervised learning,''
  {\em arXiv preprint arXiv:1610.02242}, 2016.

\bibitem{lu2019semi}
M.~Y. Lu and et~al., ``Semi-supervised histology classification using deep
  multiple instance learning and contrastive predictive coding,'' {\em arXiv
  preprint arXiv:1910.10825}, 2019.

\bibitem{jaiswal2019semi}
A.~K. Jaiswal and et~al., ``Semi-supervised learning for cancer detection of
  lymph node metastases,'' {\em arXiv preprint arXiv:1906.09587}, 2019.

\bibitem{su2019local}
H.~Su and et~al., ``Local and global consistency regularized mean teacher for
  semi-supervised nuclei classification,'' in {\em International Conference on
  Medical Image Computing and Computer-Assisted Intervention}, pp.~559--567,
  Springer, 2019.

\bibitem{shaw2020teacher}
S.~Shaw, M.~Pajak, A.~Lisowska, S.~A. Tsaftaris, and A.~Q. O'Neil,
  ``Teacher-student chain for efficient semi-supervised histology image
  classification,'' {\em arXiv preprint arXiv:2003.08797}, 2020.

\bibitem{tzeng2017adversarial}
E.~Tzeng, J.~Hoffman, K.~Saenko, and T.~Darrell, ``Adversarial discriminative
  domain adaptation,'' in {\em Proceedings of the IEEE conference on computer
  vision and pattern recognition}, pp.~7167--7176, 2017.

\bibitem{ganin2016domain}
Y.~Ganin, E.~Ustinova, H.~Ajakan, P.~Germain, H.~Larochelle, F.~Laviolette,
  M.~Marchand, and V.~Lempitsky, ``Domain-adversarial training of neural
  networks,'' {\em The Journal of Machine Learning Research}, vol.~17, no.~1,
  pp.~2096--2030, 2016.

\bibitem{shen2018wasserstein}
J.~Shen, Y.~Qu, W.~Zhang, and Y.~Yu, ``Wasserstein distance guided
  representation learning for domain adaptation,'' in {\em Thirty-Second AAAI
  Conference on Artificial Intelligence}, 2018.

\bibitem{bousmalis2017unsupervised}
K.~Bousmalis, N.~Silberman, D.~Dohan, D.~Erhan, and D.~Krishnan, ``Unsupervised
  pixel-level domain adaptation with generative adversarial networks,'' in {\em
  Proceedings of the IEEE conference on computer vision and pattern
  recognition}, pp.~3722--3731, 2017.

\bibitem{ren2018adversarial}
J.~Ren, I.~Hacihaliloglu, E.~A. Singer, D.~J. Foran, and X.~Qi, ``Adversarial
  domain adaptation for classification of prostate histopathology whole-slide
  images,'' in {\em International Conference on Medical Image Computing and
  Computer-Assisted Intervention}, pp.~201--209, Springer, 2018.

\bibitem{xing2019adversarial}
F.~Xing, T.~Bennett, and D.~Ghosh, ``Adversarial domain adaptation and
  pseudo-labeling for cross-modality microscopy image quantification,'' in {\em
  International Conference on Medical Image Computing and Computer-Assisted
  Intervention}, pp.~740--749, Springer, 2019.

\bibitem{stacke2019closer}
K.~Stacke, G.~Eilertsen, J.~Unger, and C.~Lundstr{\"o}m, ``A closer look at
  domain shift for deep learning in histopathology,'' {\em arXiv preprint
  arXiv:1909.11575}, 2019.

\bibitem{hinton2006reducing}
G.~E. Hinton and R.~R. Salakhutdinov, ``Reducing the dimensionality of data
  with neural networks,'' {\em science}, vol.~313, no.~5786, pp.~504--507,
  2006.

\bibitem{pathak2016context}
D.~Pathak, P.~Krahenbuhl, J.~Donahue, T.~Darrell, and A.~A. Efros, ``Context
  encoders: Feature learning by inpainting,'' in {\em Proceedings of the IEEE
  conference on computer vision and pattern recognition}, pp.~2536--2544, 2016.

\bibitem{larsson2017colorization}
G.~Larsson, M.~Maire, and G.~Shakhnarovich, ``Colorization as a proxy task for
  visual understanding,'' in {\em Proceedings of the IEEE Conference on
  Computer Vision and Pattern Recognition}, pp.~6874--6883, 2017.

\bibitem{gidaris2018unsupervised}
S.~Gidaris, P.~Singh, and N.~Komodakis, ``Unsupervised representation learning
  by predicting image rotations,'' {\em arXiv preprint arXiv:1803.07728}, 2018.

\bibitem{noroozi2017representation}
M.~Noroozi, H.~Pirsiavash, and P.~Favaro, ``Representation learning by learning
  to count,'' in {\em Proceedings of the IEEE International Conference on
  Computer Vision}, pp.~5898--5906, 2017.

\bibitem{henaff2019data}
O.~J. H{\'e}naff, A.~Srinivas, J.~De~Fauw, A.~Razavi, C.~Doersch, S.~Eslami,
  and A.~v.~d. Oord, ``Data-efficient image recognition with contrastive
  predictive coding,'' {\em arXiv preprint arXiv:1905.09272}, 2019.

\bibitem{dosovitskiy2014discriminative}
A.~Dosovitskiy, J.~T. Springenberg, M.~Riedmiller, and T.~Brox,
  ``Discriminative unsupervised feature learning with convolutional neural
  networks,'' in {\em Advances in neural information processing systems},
  pp.~766--774, 2014.

\bibitem{gildenblat2019self}
J.~Gildenblat and E.~Klaiman, ``Self-supervised similarity learning for digital
  pathology,'' {\em arXiv preprint arXiv:1905.08139}, 2019.

\bibitem{freeman1964apictorial}
H.~Freeman and L.~Garder, ``Apictorial jigsaw puzzles: The computer solution of
  a problem in pattern recognition,'' {\em IEEE Transactions on Electronic
  Computers}, no.~2, pp.~118--127, 1964.

\bibitem{ruifrok2001quantification}
A.~C. Ruifrok, D.~A. Johnston, {\em et~al.}, ``Quantification of histochemical
  staining by color deconvolution,'' {\em Analytical and quantitative cytology
  and histology}, vol.~23, no.~4, pp.~291--299, 2001.

\bibitem{salimans2016improved}
T.~Salimans, I.~Goodfellow, W.~Zaremba, V.~Cheung, A.~Radford, and X.~Chen,
  ``Improved techniques for training gans,'' in {\em Advances in neural
  information processing systems}, pp.~2234--2242, 2016.

\bibitem{he2016deep}
K.~He, X.~Zhang, S.~Ren, and J.~Sun, ``Deep residual learning for image
  recognition,'' in {\em Proceedings of the IEEE conference on computer vision
  and pattern recognition}, pp.~770--778, 2016.

\bibitem{ronneberger2015u}
O.~Ronneberger, P.~Fischer, and T.~Brox, ``U-net: Convolutional networks for
  biomedical image segmentation,'' in {\em International Conference on Medical
  image computing and computer-assisted intervention}, pp.~234--241, Springer,
  2015.

\end{thebibliography}

\end{document}